\documentclass[conference]{IEEEtran}
\IEEEoverridecommandlockouts
\usepackage{cite}
\usepackage{amsmath,amssymb,amsfonts}
\usepackage{algorithmic}
\usepackage{graphicx}
\usepackage{textcomp}
\usepackage{xcolor}
\usepackage{comment}
\usepackage{color}
\def\BibTeX{{\rm B\kern-.05em{\sc i\kern-.025em b}\kern-.08em
    T\kern-.1667em\lower.7ex\hbox{E}\kern-.125emX}}

\definecolor{red}{rgb}{0.95,0.4,0.4}
\definecolor{purered}{rgb}{1,0,0}
\definecolor{blue}{rgb}{0.4,0.4,0.95}
\definecolor{darkblue}{rgb}{0,0,0.8}
\definecolor{darkred}{rgb}{1,0,0}
\definecolor{darkgreen}{rgb}{0,0.5,0}
\definecolor{grey}{rgb}{0.6,0.6,0.6}
\definecolor{col1}{RGB}{232, 161, 148}
\definecolor{col2}{RGB}{148, 187, 232}
\definecolor{lightgrey}{rgb}{0.85,0.85,0.85}
\definecolor{lightlightgrey}{rgb}{0.9,0.9,0.9}
\definecolor{verylightBG}{rgb}{0.9,0.99,0.99}
\definecolor{darkgreen}{rgb}{0.3, 0.75, 0.3}
\definecolor{darkgrey}{rgb}{0.8,0.35,0.35}

\newcommand\blfootnote[1]{%
  \begingroup
  \renewcommand\thefootnote{}\footnote{#1}%
  \addtocounter{footnote}{-1}%
  \endgroup
}

\begin{document}

\title{A Brief Survey on Person Recognition at a Distance}
\author{\IEEEauthorblockN{Chrisopher B. Nalty*, Neehar Peri*, Joshua Gleason*, \\ Carlos D. Castillo, Shuowen Hu, Thirimachos Bourlai, Rama Chellappa}}

\maketitle

\begin{abstract}
Person recognition at a distance entails recognizing the identity of an individual appearing in images or videos collected by long-range imaging systems such as drones or surveillance cameras. Despite recent advances in deep convolutional neural networks (DCNNs), this remains challenging. Images or videos collected by long-range cameras often suffer from atmospheric turbulence, blur, low-resolution, unconstrained poses, and poor illumination. In this paper, we provide a brief survey of recent advances in person recognition at a distance. In particular, we review recent work in multi-spectral face verification, person re-identification, and gait-based analysis techniques. Furthermore, we discuss the merits and drawbacks of existing approaches and identify important, yet under explored challenges for deploying remote person recognition systems in-the-wild.
\end{abstract}

\begin{IEEEkeywords}
Person Recognition, Person Re-Identification, Person Verification, Long Range Recognition, Person Recognition Survey, Biometrics
\end{IEEEkeywords}

\blfootnote{* First three authors contributed equally.}

\section{Introduction}
\begin{figure}[t]
    \centering
    \includegraphics[trim=0cm 15cm 1cm 0cm, clip, width=\linewidth]{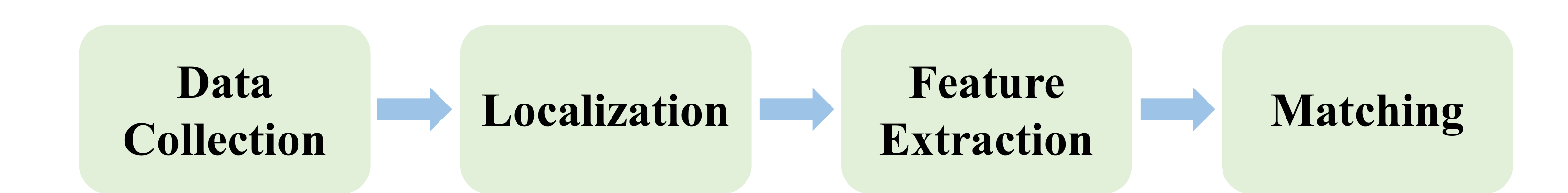}
    \caption{The person recognition pipeline contains four steps. First, we acquire an image or video of a person-of-interest. Second, we detect a bounding box around the subject. Next, we we extract deep features that can be used to verify the subject's identity. Lastly, we compare this deep feature against a large gallery for verification. Recent work focuses on improving the discriminative quality of the feature extractor using novel architectures and training techniques.}
    \label{fig:pipeline}
    \vspace{-9pt}
\end{figure}

Person recognition involves extracting features to identify or verify a person's identity. Although historical approaches used hand crafted features for recognition \cite{eigenface, belhumeur_fisher_eigen}, large-scale annotated datasets and advances in deep convolutional neural networks (DCNNs) have yielded significant improvements in person recognition. State-of-the-art person recognition algorithms can match the performance of forensic experts \cite{phillips2018_forensic} and are commonly used in real-world applications including missing person identification, contact-less authentication, and intelligent analytics in smart cities \cite{chellappa1, chellappa1995human, tome_applications}. Recent work has focused on developing novel architectures, representations, and training techniques for person recognition at a distance. Most person recognition algorithms follow a four-stage pipeline, described in Figure \ref{fig:pipeline}: acquiring an image or video of the person-of-interest, detecting and localizing the person, generating a feature descriptor, and classifying the descriptor into one of several known subjects for recognition or comparing it to a gallery of potential subjects for verification. This general template is used for tasks such as face verification, person re-identification, and gait analysis.

\textbf{Face Verification.} 
Face verification is the task of determining whether two face images belong to the same person. In unconstrained settings, face verification can be difficult due to variations in viewpoint, low-resolution, occlusion, and image quality. Imbalanced data quality in training sets can also hinder performance. For example, models trained on data biased towards frontal-faces will perform poorly on profile-face recognition. Similarly, models trained only on light-skinned individuals will have higher error rates for dark-skinned individuals \cite{face_bias}. These biases make it challenging to deploy face verification algorithms in-the-wild.

\textbf{Person Re-Identification.} 
The goal of person re-identification (Re-ID) is to determine whether a person-of-interest has appeared in another place at a different time, using a query image, video sequence, or text description \cite{ye_deep_reid_survey}. Person re-id is similar to face verification, but typically involves training with and evaluating on significantly fewer identities. This problem is challenging because it often requires re-identifying a person across non-overlapping cameras. Successful algorithms must be robust to occlusion, changes in pose, changes in lighting, and cluttered backgrounds.  Person re-identification is often used in applications where face verification is not possible due to low resolution or long distance applications. Since people may wear the same color clothes, or have similar height and body types, person re-identification is often less accurate than face verification. 

\textbf{Gait Analysis.}
Gait analysis is a technique for recognizing people based on their pose and motion. The standard pipeline for gait analysis involves extracting gait features from a sequence of images corresponding to a person. These features are typically extracted over a gait cycle, which is defined as the sequence of movements between two consecutive contacts of the same foot with the ground. Gait analysis is considered a soft biometric, a relatively weak discriminative feature for person recognition \cite{isaac_gait}. Similar to person re-identification, gait analysis can be performed from a distance. 

\textbf{Contributions.} In this paper, we provide an overview of person recognition methods, including person re-identification, face verification, and gait analysis for real-world conditions. We discuss the problem setup, evaluation metrics, standard datasets, and representative methods and conclude by examining the challenges of deploying person recognition systems in real-world environments.

\section{Face Verification}
\begin{figure*}[ht]
    \centering
    \includegraphics[trim=0cm 10cm 1cm 1cm, clip, width=0.95\linewidth]{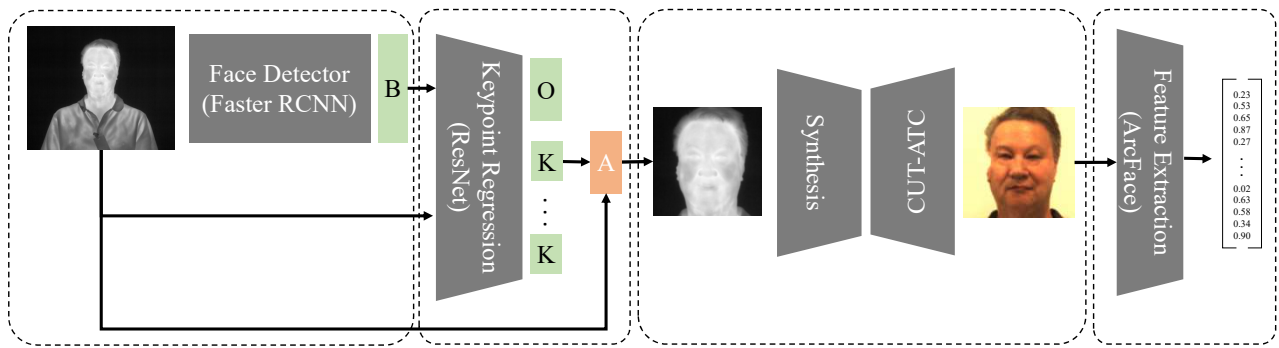}
   \vskip-10pt\caption{The face verification pipeline contains four steps. First, we detect a bounding box around the subject's face. Next, we use the coarse-grained bounding box (Output B) to regress keypoint locations (Output K). We use these estimated landmarks to align the input image (Function A). Optionally, if the input image is from a different domain, we use the aligned face image and synthesize a visible spectrum output. Lastly, we extract deep features using an off-the-shelf model that can be used to verify the subject's identity. Figure taken from \cite{peri21_thermal2visible}.}
    \label{fig:t2v_pipeline}
    \vspace{-9pt}

\end{figure*}

In this section, we describe the face verification problem setup and highlight several contemporary approaches. 

\textbf{Problem Setup.}
Face verification involves determining if two face images belong to the same person. Advances in deep convolutional neural networks (DCNNs) and large visible-spectrum datasets have led to significant improvements in face verification. However, existing approaches still struggle to match faces with large pose variations, at a distance, and in low-light or nighttime conditions. There are two main ways to train DCNNs for face recognition: training a multi-class classifier that can separate different identities in the training set using a softmax classifier, or learning an embedding directly using a triplet loss. Both approaches have been shown to produce state-of-the-art performance on face recognition tasks when trained on large datasets.

\textbf{Evaluation Metrics.} Three evaluation metrics are commonly used for face verification, including true acceptance rate (TAR), false acceptance rate (FAR), and equal error rate (EER) \cite{wang2021deep}. TAR measures the rate a system correctly verifies a true claim of identity. FAR is the rate a system incorrectly verifies a claim of identity. EER describes the point where the TAR and FAR are equal.

\textbf{Datasets.}
The VGGFace dataset \cite{parkhi15_vggface} is a large-scale dataset designed for face recognition that contains 2.6 million images of 2,622 people. VGGFace is not very diverse in terms of pose, with 95\% of the images being frontal and only 5\% being non-frontal. The Megaface challenge \cite{kemelmacher2016megaface} is a recent face recognition benchmark that includes a gallery with one million images of one million people. Megaface uses the Facescrub dataset \cite{facescrub} as the query set. The CASIA WebFace dataset \cite{casiawebface} includes 494,414 images of 10,575 people, but does not provide labels. The CelebFaces+ dataset contains 10,177 subjects and 202,599 images. CelebA \cite{liu2015faceattributes} added five landmark locations and 40 binary attributes to the CelebFaces+ dataset. UMDFaces \cite{bansal2016umdfaces} includes 367,888 face annotations for 8,277 subjects, and provides human-curated bounding boxes for faces, as well as estimated pose, keypoints, and apparent gender from a DCNN model. WebFace260M \cite{zhu2021webface260m} is the largest currently available public dataset, containing noisy annotations for four million identities and 260 million faces and clean annotations for two million identities and 42 million images.

Several thermal-to-visible datasets have been released to facilitate low-light and nighttime face verification. The UND dataset \cite{chen03_und} contains 241 identities with four low-resolution images each. The NVIE dataset \cite{wang10_nvie} captures subjects with a wide range of expressions, with and without glasses. MMFD \cite{hu16_mmfd} provides synchronized imagery of visible, LWIR, and Polarimetric LWIR data at variable distances from the camera. ULFMT \cite{ghiass18_ulfmtv} contains unsynchronized MWIR and visible video recordings of 238 subjects captured under variable conditions. The Tufts Face Database \cite{panetta20_tufts} contains LWIR, NIR, 2D, computer sketches and 3D point clouds for 100 subjects. The ARL-VTF dataset \cite{poster21_arlvtf} is a recently released large-scale LWIR-visible paired face dataset containing 395 subjects with variable pose and expression along with keypoint annotations. Peri et. al. \cite{peri21_thermal2visible} presents MILAB-VTF(B), a challenging multi-spectral face dataset of indoor and long-range outdoor thermal-visible face imagery from 400 subjects.

\textbf{Visible-to-Visible Verification.}
Deep face recognition involves training a CNN-based classifier on a large dataset of faces and comparing the deep features from the penultimate layer of the trained model. This comparison is typically done using cosine distance. A notable early example of this approach is  VGGFace \cite{parkhi15_vggface}, which uses a DCNN architecture trained for classification using cross-entropy loss, followed by an additional triplet embedding function to improve the discriminative characteristics of the deep features. Various augmentations to cross-entropy loss can yield improved performance. For example, Crystal loss \cite{ranjan17_l2softmax} uses an $\ell_2$ normalized softmax to optimize the angular margin and encourage features that are more discriminative with respect to cosine similarity. Other works have proposed similar augmentations to cross-entropy loss to improve margin characteristics. CosFace \cite{cosface} disables the bias term from the final fully-connected layer and proposes using cosine similarity with an additive penalty in place of the inner product used in cross-entropy loss to improve angular margin. ArcFace \cite{deng19_arcface} builds on CosFace by proposing an additional additive penalty to the angular margin directly. MagFace \cite{magface} further builds on ArcFace by introducing dynamic margins based on the hardness of face images. Finally, Partial-FC \cite{partialfc} proposes a method for training state-of-the-art models using margin-based losses on previously intractably large datasets by back-propagating only a fraction of the gradients corresponding to negative centroids in the classification layer. Lastly, \cite{dosanddonts} proposes best practices and empirically answers common questions that arise when building face recognition systems. 

Several works focus on the sub-problem problem of profile-to-frontal matching. Recent works use a conditional generative adversarial network to synthesize a frontal-view of a face based on its profile-view. TP-GAN \cite{huang17_tpgan} proposed a dual-path generator that concatenates a coarsely generated frontal face with local profile facial features to generate a high-quality frontal view. \cite{zhao18_pim} extended TP-GAN to jointly learn frontal face generation and discriminative feature embeddings for end-to-end face verification. \cite{li19_m2fpa, yin20_dagan} used attention-guided synthesis with part masks to frontalize profile face images. More recently, \cite{di21_dalgan} proposed a contrastive learning approach for frontalization, achieving strong performance without using additional part annotations.

\textbf{Thermal-to-Visible Verification.}
Similar to profile-to-frontal matching, recent works address low-light and nighttime face verification using a conditional generative adversarial network to synthesize visible-spectrum faces given thermal-spectrum input. This facilitates reusing existing deep face recognition systems without fine-tuning or re-training. GAN-VFS \cite{he17_ganvfs} proposed an encoder-decoder structure that directly translates polarimetric images to the visible domain while enforcing a perceptual loss on intermediate features so that they closely resemble the intermediate feature embedding from a fined-tuned VGG-16 feature extractor. Similarly, \cite{di19_selfattn} extended CycleGAN to enforce an identity loss to ensure features from synthesized images are close to the corresponding real image features and proposed a feature fusion of both polarimetric and visible image features to improve verification robustness. Unlike most synthesis-based methods, \cite{fondje20_crossdomain} directly adapted intermediate features from both thermal and visible images using truncated fixed feature extractors to learn a domain invariant representation for cross-domain matching. More recently, \cite{di21_attributeguided} introduced a method that incorporates facial attributes by pooling latent features with attribute features and synthesized visible domain images at multiple scales to guide face synthesis effectively. Similarly, \cite{mallat19_cascade} exploited multi-scale information for higher resolution generation with less training data using a series of cascade refinement networks. \cite{peri21_thermal2visible} studies the impact of keypoint-based face alignment, pixel-level correspondence, and feature-level identity classification, adapting off-the-shelf domain adaptation algorithms to achieve state-of-the-art results. We present the end-to-end inference pipeline in Figure \ref{fig:t2v_pipeline}

\textbf{Environmental Degradation.}
Long-range face verification must contend with environmental degradation and nuisance factors not readily apparent in indoor and close-range verification. Two approaches are commonly used for restoring images degraded by atmospheric turbulence. ``Lucky imaging'' \cite{lucky1, lucky2} involves selecting and combining frames from a turbulence-degraded video that are of higher quality. Registration-fusion \cite{hirsch2010efficient, zhu2012removing} involves constructing a reference image and aligning distorted frames using non-rigid image registration. After alignment, the registered images are combined and a restoration algorithm is applied to deblur the result. While these methods have been successful, they assume that the input frames contain static scenes, which is not always the case. Recently, several CNN-based face restoration algorithms have proposed using GANs to deblur face images \cite{conv_gan}. However, these methods are not effective for dealing with degradation caused by turbulence, which can include motion blur, out-of-focus blur, and compression artifacts. \cite{atfacegan} proposes a two-path model that constructs a clean texture-preserving face image by simultaneously disentangling blur and deformation.

\section{Person Re-Identification}
In this section, we describe the typical problem setup for person re-identification and briefly highlight several contemporary approaches. 

\textbf{Problem Setup.}
Person re-identification was originally introduced by \cite{WANG20133} as a method for improving multi-camera tracking, combining appearance features with traditional tracking algorithms \cite{wojke2017deepsort, peri2020towards, khorramshahi2019attention}. However, it is now considered a standalone task and is evaluated using independent benchmarks. These benchmarks assume that images are taken from non-overlapping cameras, meaning that images will have different standoff distances, different image quality, and may capture the subject in different poses. For evaluation under a closed-world setup, benchmarks assume that every identity in the query set has a matching pair in the gallery. 

\textbf{Evaluation Metrics.} Two metrics used to evaluate the performance of person re-identification system are Rank-k matching accuracy and mean average precision. Rank-k matching accuracy measures the probability that a correct match will appear in the top $k$ ranked results and assumes that there is only one matching image in the gallery set. Mean average precision, which is commonly used in information retrieval, is a more comprehensive metric that takes into account multiple matches in a gallery and provides a more accurate measure of overall performance \cite{ye_deep_reid_survey}.

\textbf{Datasets.} We briefly highlight five popular person re-id datasets. Two datasets are image-based, while the remaining three are video-based. The image-based datasets are Market-1501 \cite{zheng2015scalable}, which contains 1501 identities captured by six cameras, and CUHK03  \cite{li2014deepreid}, which contains 14,097 images of 1,467 different identities. The three video-based datasets are PRID2011  \cite{prid2011}, which contains 400 sequences of 200 people, iLIDS-VID \cite{iLIDS-VID}, which contains 600 sequences of 300 subjects, and MARS \cite{zheng2016mars}, which is the largest of the three datasets with 1,261 identities and around 20,000 videos.

\begin{figure}
    \centering
    \includegraphics[width=\linewidth]{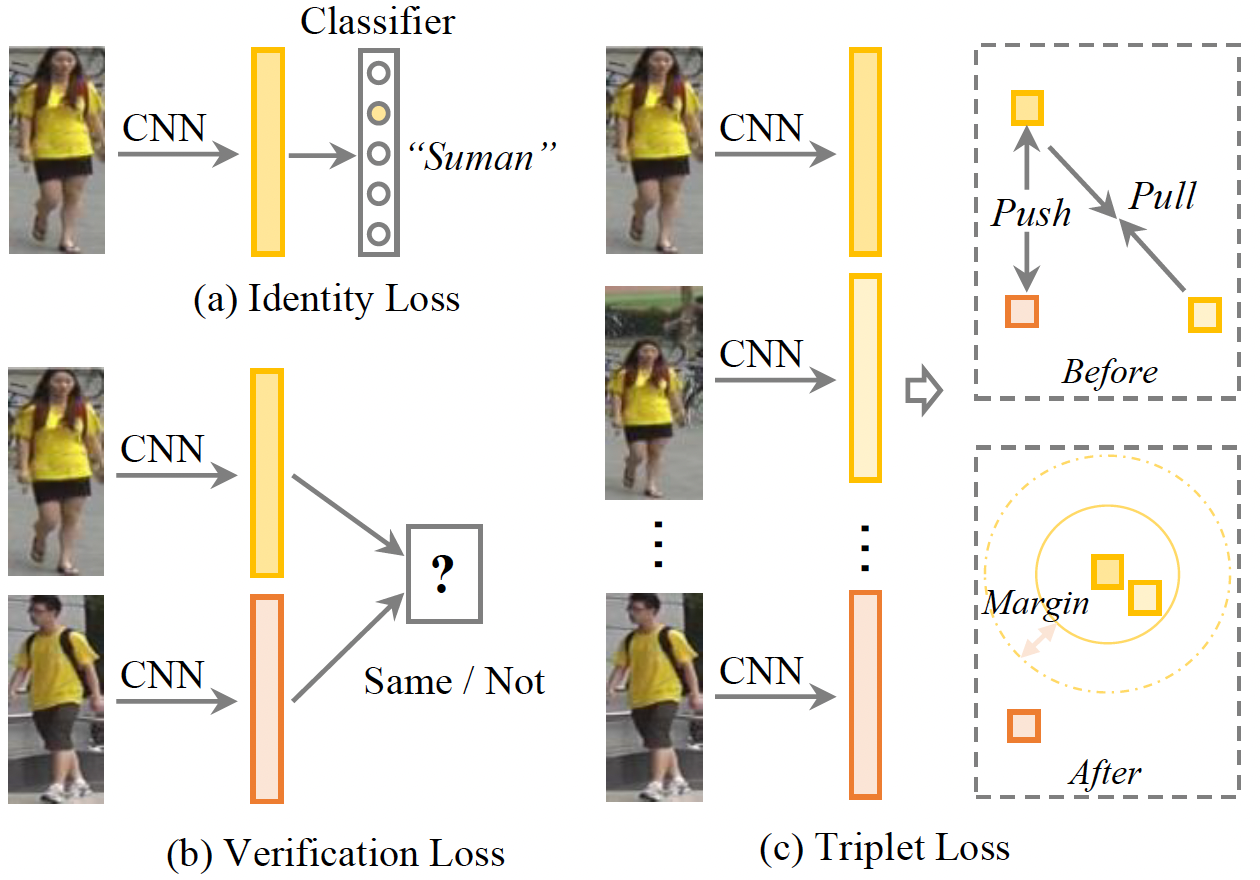}
    \caption{Identity loss and triplet loss are widely used for training person re-identification systems. Specifically, (a) identity loss simply trains a softmax classifier over the total number of identities in the training set to learn discriminative features, while (b) triplet loss selects an anchor, positive, and negative example such that the distance between the anchor and positive example are less than the anchor and negative example. identity loss and triplet loss are often used together, as this has been shown to improve performance. Figure taken from \cite{ye_deep_reid_survey}.}
    \label{fig:reid_loss}
    \vspace{-9pt}

\end{figure}

\textbf{Representation Learning.}
Person re-identification methods can be divided into global and local feature learning approaches. Global feature learning aims to learn a single feature that represents the entire image, while local feature learning segments the body into coarse regions or keypoints to extract locality-specific features. Person re-id methods, and representation learning techniques in general, are also widely applicable for object re-identification tasks \cite{khorramshahi2019dual, khorramshahi2020devil}.

\textit{Global Features.} Global feature representation learning is widely adopted, and serves as a simple, yet strong baseline. Identity loss \cite{zheng2017person} is widely used in the Re-ID community because it treats each identity as a distinct class and trains the model using a multi-class classification loss. Chung et. al. \cite{chung2017two} uses a two stream Siamese network to capture both RGB information and temporal information through optical flow. Luo et. al. \cite{luo19_bagoftricks} proposes a strong baseline using a ``bag of training tricks'', including warm-up learning rate, random erasing, label smoothing, and center loss. 

Other approaches, such as multi-scale deep representation learning \cite{li2015multi, qian2019leader} and attention \cite{yang2019attention} are widely used. Pixel-level attention \cite{li2018harmonious}, channel-wise feature response re-weighting \cite{wang2018mancs}, and background suppression \cite{shen2018end} have also been studied. Additionally, context-aware attentive feature learning methods \cite{si2018dual} incorporate both intra-sequence and inter-sequence attention for pair-wise feature alignment and refinement. Group similarity \cite{chen2018group} is another approach that leverages cross-image attention. These methods can improve robustness against imperfect detection, as well as enhance feature learning by mining relations across multiple images. 

\newpage

\textit{Local Features.} Local feature representation learning focuses on learning part or region-aggregated features. Semantically meaningful body part regions are typically automatically generated using human pose estimation \cite{openpose} networks, or are roughly divided horizontally. A common approach for automatic body part detection include combining a global representation with local part features, using techniques such as multi-channel aggregation \cite{cheng2016person}, multi-scale context-aware convolutions \cite{li2017learning}, multistage feature decomposition \cite{zhao2017spindle}, and bilinear-pooling \cite{suh2018part}. Approaches that use pose estimation to obtain semantically meaningful body parts provide well-aligned part features but may require an additional pose detector and are susceptible to noisy pose detections. Another popular solution for enhancing robustness against background clutter is to use pose-driven matching \cite{sun2018beyond}, pose-guided part attention modules \cite{xu2018attention}, and semantically-aligned parts \cite{zhang2019densely}. Horizontally-divided regions are more flexible, and don't require using a pose estimation network, but are less accurate due to perspective distortions at longer standoff distances.

\textbf{Metric Learning.}
Metric learning optimizes the relationship between pairs of data points, either using a contrastive loss \cite{varior2016siamese} or margin ranking loss. Triplet loss \cite{hermans2017defense} ensures that the $\ell_2$ distance between positive pairs should be smaller than the distance between negative pairs by a fixed margin. However, directly using this loss function can result in learning feature with limited discrimination because a large proportion of easy triplets will dominate the training process. To address this issue, various methods for selecting informative triplets have been developed \cite{shi2016embedding, yu2018hard}. These methods aim to find semi-hard triples and have been shown to improve the discriminative ability of re-identification models. Some methods also use a point-to-set similarity strategy to enhance robustness against outlier samples \cite{zhou2017point}. Additionally, a quadruplet deep network\cite{chen2017beyond} has been proposed, which uses margin-based online hard negative mining to optimize the quadruplet relationship between an anchor, one positive, and two negative samples. The quadruplet loss has been shown to improve both intra-class and inter-class variation. The combination of triplet loss and identity loss (as shown in Figure \ref{fig:reid_loss}), is a popular solution for training deep re-identification models.

\textbf{Ranking Optimization.}
Ranking optimization is an important post-processing technique used to improve retrieval performance. This entails taking an initial ranked gallery and optimizing the ranking order, either through automatic gallery-to-gallery similarity mining \cite{ye2015ranking} or human-in-the-loop labeling \cite{liu2013pop}. Re-ranking aims to use gallery-to-gallery similarity to improve the initial ranking list using easy matches in the gallery to expand the query set to find harder matches in the gallery. The widely-used k-reciprocal re-ranking method \cite{zhong2017re} mines contextual information to do this. Other methods, such as \cite{bai2017scalable} take a geometric structure-based approach. Local blurring re-ranking \cite{sarfraz2018pose} uses a clustering structure to improve neighborhood similarity measurement. 

\section{Gait Analysis}
In this section, we discuss the problem setup and datasets use for gait analysis, as well as representative methods. 

\textbf{Problem Setup.} Gait analysis for person identification is concerned with recognizing individuals from their gait. It was pioneered by the medical community and has been adopted as a form of soft biometrics. Importantly, gait analysis can be performed at a distance \cite{shen2022comprehensive}. However, it is not as effective as person re-identification or face verification. Gait recognition can be divided into two categories: model-based and model-free. Model-free methods tend to be simpler, more efficient and more commonly used compared to model-based methods.

\textbf{Evaluation Metrics.}
Gait recognition evaluation follows many of the same metrics as face verification, including true acceptance rate, false acceptance rate and equal error rate.

\textbf{Datasets.} Many datasets have been introduced to facilitate work in gait analysis. The largest dataset of the SOTON \cite{soton} database contains 115 subjects and over 2000 sequences from indoor and outdoor conditions and includes both silhouette and RGB images. The USF HumanID \cite{humanID} database contains 122 subjects and nearly 2,000 outdoor RGB sequences, including examples with significant occlusion (e.g. individuals carrying briefcases). CASIA \cite{casia_gait} is the most popular gait recongition database, containing five datasets with RGB, infrared, and silhouette information in both indoor and outdoor settings. OU-ISIR \cite{ou-isir} is also composed of multiple databases, and captures gait demonstrations in both indoor and outdoor settings. The largest OU-ISIR dataset annotates over 63k subjects. The TUM-GAID dataset \cite{tum-gaid} includes 300 subjects walking indoors, including examples of individuals carrying backpacks, and provides RGB, depth and audio information. OU-MVLP \cite{An_TBIOM_OUMVLP_Pose} is a large database including 10,000 subjects and over 250,000 sequences. BRIAR \cite{cornett2022expanding} is a recently introduced dataset with 350,000 still images and over 1,300 hours of video footage of approximately 1,000 subjects.

\textbf{Gait Representations.}
Gait frames are often represented through silhouette images and skeletal representations. Silhouettes are binary images typically created by background subtraction, but can also be created using DCNN based human segmentation models \cite{openpose}. Silhouettes are more affected than skeletal models by changes in clothing and camera angle. Skeletal models require accurate DCNNs for real-world identification. Until recently, pose estimation methods were not accurate enough to reliably generate skeletal models. Recent works have started to use skeletal information instead of silhouettes. Skeletal modeling provide more reliable features than silhouettes, because skeletal representations should always be the same regardless of an individual's clothing and should be less affected by factors such as viewing angle. 

\begin{figure}
    \centering
    \includegraphics[width=\linewidth]{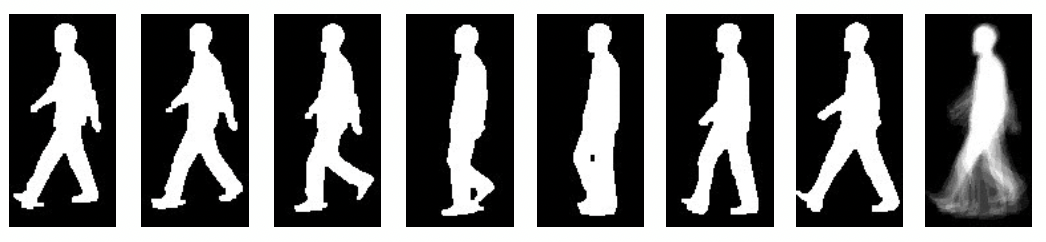}
    \caption{Silhouettes are often used to represent gait, but are less accurate than skeletal models. Silhouettes are not robust to temporary clothing deformations or occlusions. The last image shown above is the Gait Energy Image (GEI) aggregated from the first seven frames (on the left). Figure taken from \cite{isaac_gait}.}
    \label{fig:gait_representation}
    \vspace{-9pt}
\end{figure}

In addition to single-frame representations, gait can also represented by multi-frame sequences. There are three main types of temporal representations used in gait recognition. The most common are Gait Energy Images (GEIs), which are aggregates of silhouette frames summed into a single image (see Figure \ref{fig:gait_representation}). The second type are convolution templates, in which each image in a sequence is processed through a convolution and pooling layer, then aggregated into a template. The last representation is sequence volumes, which are simply stacks of the gait images into a single multi-channel input image for a neural network. Sequence volumes have been growing in popularity over traditional methods. 

\textbf{Model-Based Approaches.}
Model-based methods use a deformable model of the human body or motion to make inferences. This increases the computational complexity of the problem and generally requires higher-quality imaging. However, because the body is modeled by a 2D skeleton or 3D mesh, these methods are often more accurate when dealing with occlusions. A recent example of model-based pose estimation is PoseGait \cite{posegait} which extracts deep features directly from a 3D pose estimate. ModelGait \cite{modelgait_2020_ACCV} proposes an end-to-end system that estimates pose using a CNN which is then used to predict identity. MvModelGait \cite{mvmodelgait} extends ModelGait by incorporating multi-view information during training while supporting single-view at test time. Finally, \cite{gaitgraph} presents a graph convolutional network approach that encodes sequences of poses into a single identity descriptor.

\textbf{Model-Free Approaches.}
Model-free or appearance-based methods only use image information to make inferences on sequences. These methods work well with low-quality imaging but can struggle with occlusions. Some recent methods include multi-scale temporal 3D CNNs \cite{context_sensitive_gait2021, 3dconvgait_ICCV2021}, which are among the highest-performing. \cite{crossview_TIP2020} uses a temporal attention mechanism to produce discriminative cross-view representations. GaitGL \cite{lin2022gaitgl} presents a strong deep learning based approach that uses deep learning to permit discriminative features at both the global and local scale. Results using GaitGL on the recently introduced BRIAR dataset are presented in \cite{guo2022multi}.

\section{Conclusion}
In this paper, we provide an overview of person recognition methods, specifically, person re-identification, face verification, and gait analysis for real-world conditions. We find that current methods for person recognition suffer from data bias, which can be caused by training dataset factors such as skin tone, pose, occlusions, distance, and lighting conditions. For example, models trained for frontal face verification may not perform well when presented with profile faces, and models trained to recognize people indoors may not adapt well to outdoor scenes. Domain adaptation is a promising area of future research, but the adaptation methods used for person recognition tasks must faithfully preserve identity information.

\newpage

\bibliographystyle{IEEEtran}
\bibliography{IEEEabrv,references}

\end{document}